\pdfoutput=1

\documentclass[11pt]{article}

\usepackage{acl}

\usepackage{times}
\usepackage{latexsym}

\usepackage[T1]{fontenc}

\usepackage[utf8]{inputenc}

\usepackage{microtype}

\usepackage{amsmath}
\usepackage{amsfonts}
\usepackage{amssymb}
\usepackage{enumerate}
\usepackage{times}
\usepackage{latexsym}
\usepackage{graphicx}
\usepackage{algorithm}
\usepackage[noend]{algpseudocode}
\usepackage{amsmath}
\usepackage{booktabs}

\DeclareMathOperator*{\argmin}{arg\,min}

\usepackage{bm}
\usepackage{mathtools}
\usepackage{dsfont}

\title{Constrained Few-Shot Learning: Human-Like Low Sample Complexity Learning and Non-Episodic Text Classification}

\author{Jaron Mar \\
  The University of Auckland \\\
  \texttt{jaron.mar@auckland.ac.nz} \\\And
  Jiamou Liu \\
  The University of Auckland \\\
  \texttt{jiamou.liu@auckland.ac.nz} \\}

\begin{document}
\maketitle
\begin{abstract}

Few-shot learning (FSL) is an emergent paradigm of learning that attempts to learn to reason with low sample complexity to mimic the way humans learn, generalise and extrapolate from only a few seen examples. While FSL attempts to mimic these human characteristics, fundamentally, the task of FSL as conventionally formulated using meta-learning with episodic-based training does not in actuality align with how humans acquire and reason with knowledge. FSL with episodic training, while only requires $K$ instances of each test class, still requires a large number of labelled training instances from disjoint classes. In this paper, we introduce the novel task of constrained few-shot learning (CFSL), a special case of FSL where $M$, the number of instances of each training class is constrained such that $M \leq K$ thus applying a similar restriction during FSL training and test. We propose a method for CFSL leveraging Cat2Vec using a novel categorical contrastive loss inspired by cognitive theories such as fuzzy trace theory and prototype theory. 

\end{abstract}

\section{Introduction}



Does few-shot learning (FSL) as conventionally formulated as a \emph{$N$-way $K$-shot} learning problem really mimic the remarkable cognitive abilities humans have to learn, generalise, and extrapolate to perform reasoning? The goal of FSL is to learn a model that predicts $N$ unseen test classes using only $K$ labelled examples of each unseen test class as to mimic a humans ability to learn from sparse information. To achieve this, FSL typically requires three datasets, $\mathcal{D}_\mathrm{train} = (\mathcal{X}_\mathrm{train}$, $\mathcal{Y}_\mathrm{train})$, $\mathcal{D}_\mathrm{test} = (\mathcal{X}_\mathrm{test}$, $\mathcal{Y}_\mathrm{test})$, $\mathcal{D}_\mathrm{K} = (\mathcal{X}_\mathrm{K}$, $\mathcal{Y}_\mathrm{K})$ where $\mathcal{Y}_\mathrm{train}$ and $\mathcal{Y}_\mathrm{test}$ are disjoint and $\mathcal{D}_\mathrm{K}$ is the dataset containing $K$ labelled examples of each test class.
In practice, although the amount of labelled test data from test classes is restricted in FSL, current FSL models have only been shown to accurately predict test classes when pre-trained using large quantities of labelled training data from similar distributions \cite{finn2017model}.

This need for large quantities of training data can be attributed to task-based \emph{episodic training}, a method which makes the structure of FSL during test and training consistent \cite{vinyals2016matching} which is generally believed--although not formalised -- to learn a model that can generalise and has become synonymous with FSL. However, recent trends in FSL have seen the partial or definitive move away from episodic training due to the inefficiencies in training data exploitation compared to non-episodic based alternatives as episodes are sampling without replacement \cite{laenen2021episodes} and poor generalisation on data with distribution shifts \cite{bennequin2021bridging}.
Fundamentally, while FSL at test time mimics human decision-making, during training, the need for large quantities of training data is inconsistent with how humans learn, acquire knowledge, adapt and re-purpose existing knowledge.

We aim to make FSL training consistent with humans' ability to learn from a few examples by extrapolating using prior knowledge and only seeing a few labelled examples of each class during training and test. We thus introduce a novel sub-problem of FSL, \emph{constrained few-shot learning} (CFSL), which is FSL with the added $M$-train-shot ($M$-tshot) constraint where the number of instances of each training class is less than or equal to $M$ where we primarily focus on the extreme cases where $M=1$ and $M=5$. In such a scenario, episodic training is no longer possible as the number of training samples of each class is insufficient to construct valid episodes when sampling without replacement. 

CFSL and FSL are thematically different as CFSL does not have access to large amounts of labelled training examples from a relatively small set of classes, usually from a similar distribution to the test classes as in the conventional FSL setting. Instead, CFSL inherently promotes learning from a large set of training classes with only a few examples of each class. CFSL is therefore highly relevant for real world applications as often a few examples from a diverse set of classes are easily collected compared to many examples for a few classes, e.g., observing cars at a traffic light, you can observe multiple different year specific models of cars but observing the same year specific model multiple times is challenging.

In reality, humans often do not have multiple examples to learn from. CFSL mimics how human's can cognitively adapt and reason in this scenario and therefore has parallels with thoroughly investigated cognitive theories. For example, \emph{prototype theory}, a prevalent theory of categorisation, and \emph{fuzzy trace theory} (FTT), a dual-process theory of memory which posits how humans store information in memory and how stored information can be used to derive decisions based on two types of memory processes, gist and verbatim. We adapt Category2Vec (Cat2Vec) for CFSL, a model agnostic method for learning category embeddings inspired by FTT and levels of gist measurements \cite{stevens1946theory} which has previously been applied to the prediction of text-based risky decision-making \cite{mar2022ftt}.

\noindent \textbf{Main Contributions:} In this paper, we propose three main contributions. (1) The introduction of the novel task of constrained few-shot learning, a sub-task of FSL task that more closely resembles how humans naturally learn to extrapolate similarities between data and past experience based on a few known examples. (2) We adapt Category2Vector (Cat2Vec), a model agnostic method for CFSL inspired by cognitive theories such as prototype and fuzzy trace theory and leveraging a novel categorical contrastive loss. (3) We show that Cat2Vec outperforms baseline methods by between 5-16\% in 5-shot 1-tshot CFSL and provide unified comparisons between conventional, FSL and CFSL classifiers for text classification.

\section{Problem Formulation} \label{sec:problem formulation}
CFSL can still be formulated as a $N$-way $K$-shot learning problem where given a set of $N$ test classes, $\mathcal{Y}_\mathrm{test} = \{y_1,y_2,\ldots,y_i\}$, a test support set containing $K$ labelled examples of each test class $\mathcal{Y}_\mathrm{test}$, $S_\mathrm{test} = \{(x_i, y_i)\}^m_{i=1}$ where $m = N \times K$, the goal is to predict the classes of the test query set, $Q_\mathrm{test} = \{(x_i, y_i)\}^n_{i=1}$ where $x_i \in \mathcal{X}_\mathrm{test}$ and $y_i \in \mathcal{Y}_\mathrm{test}$. CFSL introduces a constraint on the number of examples, $M$, of each class in the training set, $\mathcal{D}_\mathrm{train}$ where $\forall_{y \in \mathcal{Y}_\mathrm{train}}\sum_{(x_i, y_i)}^{\mathcal{D}_\mathrm{train}}{\mathds{1}_{\{y = y_i\}}} \leq M$. We denote this new constraint $M$-tshot to create a $N$-way $K$-shot $M$-tshot learning problem. We focus on the case where  where $M \leq K$, in particular, the most extreme cases where $M=1$ and $M=5$.

Under this constraint, when $M$ is small, episodic training is not possible as valid episodes can no longer be sampled when sampling without replacement. Thus instead, we
combine the test support set, $S_\mathrm{test}$ and training set, $\mathcal{D}_\mathrm{train}$, that is use $\mathcal{D}_\mathrm{train}^\prime = S_\mathrm{test} \cup \mathcal{D}_\mathrm{train}$ to train a CFSL model. The goal is thus to learn a classifier that can predict $\mathcal{Y}^\prime_\mathrm{train} = \mathcal{Y}_\mathrm{train} \cup \mathcal{Y}_\mathrm{test}$, the classes in the training set as well as the $K$-shot classes, where at test time only $\mathcal{Y}_\mathrm{test}$ is considered. 


\section{Related Work}

While episodic training is not possible under CFSL, it is crucial to evaluate the advantages to understand its use and popularity in FSL. For example, models trained with episodic training are specialised towards performing specific $k$-shot tasks and varying the $K$-shots during episodic training has been formally shown to desirably minimise intra-class variance the $K$ is small and maximise intra-class variance when $K$ is large in prototypical networks \cite{cao2019theoretical}. Furthermore, reducing intra-class variation is an essential factor when using shallow meta backbone models but not critical for deeper backbones \cite{chen2019closer}. However, non-episodic alternatives do not provide a method to tune intra-class variance.

Looking beyond episodic training, recent trends in FSL have seen the partial or definitive move away from episodic training due to the inefficiencies in training data exploitation compared to non-episodic based alternatives as episodic training samples instances for query and support sets without replacement. For example, conventional classification models with self knowledge distillation perform surprisingly well on FSL \cite{tian2020rethinking}; in the same vein pretrained large language models (LMs), e.g., GTP3 without task-specific pretraining/fine-tuning are capable of text-based FSL \cite{brown2020language}; the leave-one-out training strategy as an alternative for episodic training for meta-learning \cite{chen2020closer} and; prototypical and matching networks, trained using \emph{supervised contrastive} (SC) loss \cite{khosla2020supervised}, equation \eqref{eqn:SC}, and \emph{neighbourhood component analysis} (NCA) loss, equation \eqref{eqn:NCA} \cite{laenen2021episodes}. 

These methods typically align with our approach of using $\mathcal{D}_\mathrm{train}^\prime$ as outlined in the problem formulation. While these methods can be applied to FSL, not all of these methods are applicable to CFSL for various reasons, e.g., they still rely on large amounts of labelled training data to be effective or when $M > 1$ or is small for a $M$-tshots problem, the numerator in NCA and SC losses will never be non-zero as there are no other positive instances of every class within a mini-batch.



{\small
\begin{equation} \label{eqn:SC}
    \sum_{i \in I}{\frac{-1}{\lvert P(i) \rvert}}\sum_{p \in P(i)}{\log{\frac{\exp{(v_i \cdot v_p/\tau)}}{\sum_{a \in A(i)}\exp{(v_i \cdot v_a/\tau)}}}}
\end{equation}


\begin{equation} \label{eqn:NCA}
    \sum_{i \in I}{\log{\sum_{j \in C_i}{\frac{\exp{(- \lVert Av_i - Av_j \rVert^2)}}{\sum{\exp{(- \lVert Av_i - Av_k \rVert^2)}}}}}}
\end{equation}
}

\section{Category2Vector for CFSL}


To solve the problems with current non-episodic based training methods for CFSL, we leverage Category2Vector (Cat2Vec) \cite{mar2022ftt} using a novel categorical contrastive loss which learns categorical embeddings that facilitate the process of {\em categorisation}, the act of grouping candidates into categories based on metrics such as semantic similarity. For example, figure \ref{fig:cat2vec} shows the learned category embedding of the { \tt Artist} class in the DBpedia dataset embedded near other categorical similar classes using data described in section \ref{sec:datasets}.

\begin{figure}[H]
  \centering \includegraphics[scale=0.49]{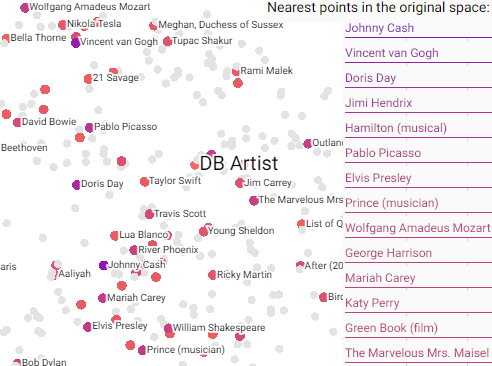}
  \caption{Cat2Vec Category Embeddings.}
 \label{fig:cat2vec}
\end{figure}

There are two prevalent cognitive theories of similarity-based categorisation, \emph{prototype} and \emph{exemplar} theory. In prototype theory, grouping is based on the similarity between candidates and \emph{prototypes}, a typical instance or representation of a category. For example,  in prototypical networks \cite{snell2017prototypical} prototypes are defined by equation \ref{eqn:prototypical}, the mean vector of embedded support points belonging to its class where $f_\phi:\mathbb{R}^D \rightarrow \mathbb{R}^O$ is an embedding function, $\phi$ are learnable parameters and $S_m$ denotes the set of examples with class $m$. In exemplar theory, grouping is based on the similarity between candidates and exemplars stored as stored memory representations, all instances of categories are stored as exemplars under exemplar theory. FSL models that are consistent with exemplar theory tend to apply pairwise comparisons between query and all support set instances which may later be compressed into a single representation by summing, pooling or averaging, e.g., relation networks \cite{sung2018learning}

\begin{equation}\label{eqn:prototypical}
    c_M = \frac{1}{|S_m|}\sum_{(x_i,y_i) \in S_m}{f_\phi(x_i)}
\end{equation}

These two cognitive theories provide a high-level explanation of categorisation, whereas FTT can provide a more fine-grained view of how information is stored in memory as gists and how reasoning applied to gist can be used for categorisation. For example, a key tenant of FTT, \emph{fuzzy processing preference} states that humans rely on the least precise gist representations necessary to make a decision. FTTs view of reasoning aligns highly and complements with prototype theory where gist representation of documents are compared to prototypes where prototypes correlate to gist representations, a single condensed representation of categories or known knowledge, to perform reasoning. While FTT aligns with prototype theory, prototypes as defined in equation \ref{eqn:prototypical} have experimentally been shown to underperform compared to current FSL methods \cite{geng2019induction}. In this paper, our view of categorisation is consistent with prototype theory; however, the way in which we define and approach prototypes is different than those defined in prototypical networks.

\subsection{Category2Vector for CFSL}\label{sec:cat2vecCFSL}

We leverage Cat2Vec to learn categorical embeddings as prototypes for CFSL that align both with prototype theory and FTT, having previously been applied to the prediction of text-based risky decision-making \cite{mar2022ftt}. A simplified Cat2Vec without categorical importance, shown in figure \ref{fig:cat2vec} is a simple yet effective model agnostic method for learning CFSL semantic category embeddings, figure \ref{fig:cat2vec}(C). Figure \ref{fig:cat2vec} shows a bi-LSTM variation of a Cat2Vec model where the encoder, \ref{fig:cat2vec}(A), can be substituted with and take advantage of any pretrained LM.

\begin{small}
\begin{equation}\label{eqn:neg_objective}
    \log \sigma \left(v_{c_{i}}\cdot v_{d_i}\right) + \sum^{N}_{j=1} \mathbb{E}_{n_j \sim P_\text{noise}(C)}[\log \sigma (-v_{n_j} \cdot v_{d_i})]
\end{equation}
\end{small}

\begin{figure}[h]
   \centering \includegraphics[scale=0.33]{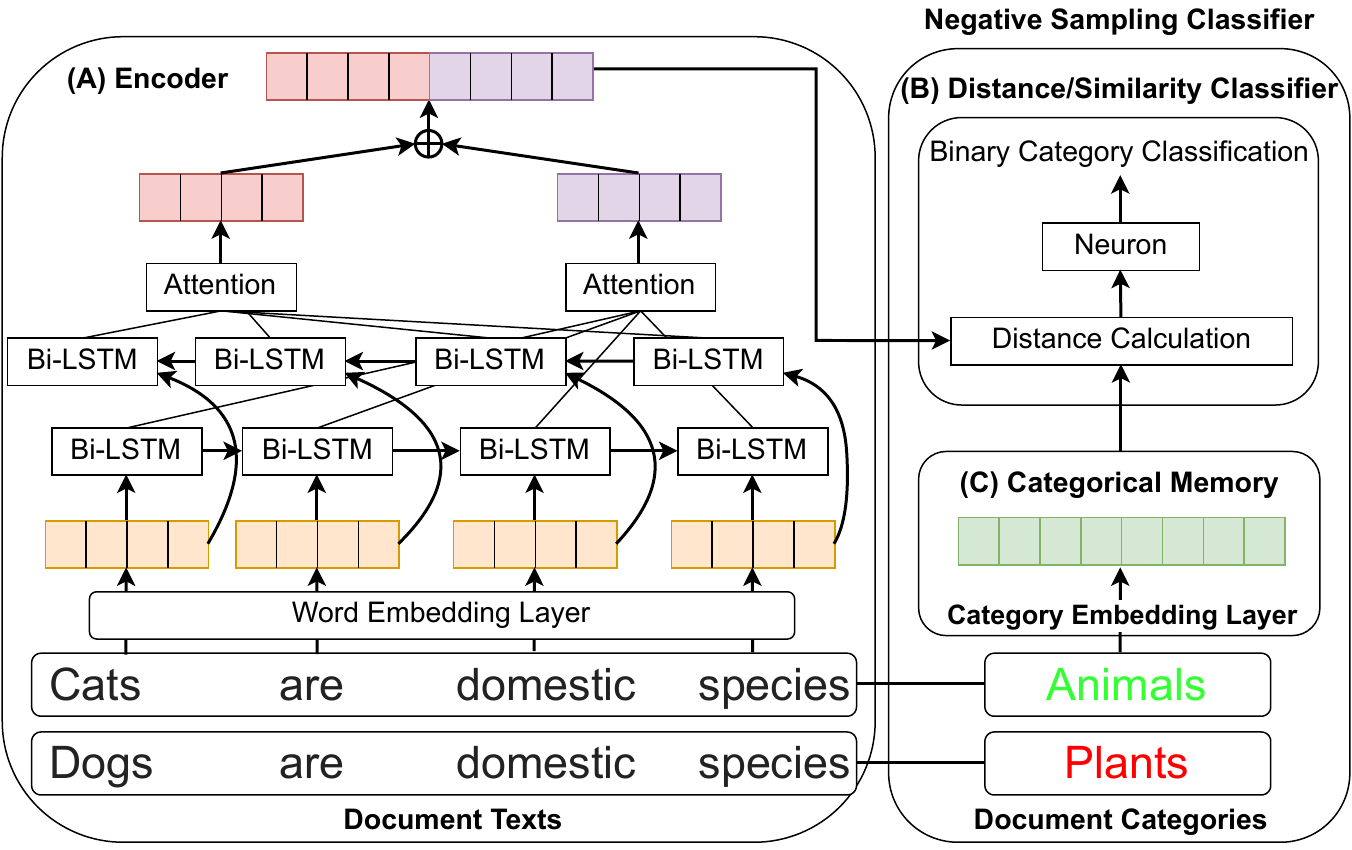}
  \caption{Bi-LSTM Cat2Vec Architecture.}
 \label{fig:cat2vec}
\end{figure}

Cat2Vec aims to maximise the average log probability that document $d_i$ belongs to category $c_i$, $\frac{1}{N}\sum^{N}_{i=1}\log P\left(c_{i} \vert d_i\right)$ using negative sampling and noise contrastive estimation (NCE) loss \cite{gutmann2012noise} given by equation \ref{eqn:neg_objective}. In this equation, $v_{d_i}$ is the encoded embedding of document $i$, $v_{c_i}$ is the learned true category embedding corresponding to document $i$ and $n_j \sim P_\text{noise}(C)$ is a noise distribution that dictates how negative document-category pairs $(d_i, n_j)$ are sampled. Cat2Vec intuitively learns to simultaneously maximises the inter-class similarity between document encodings, $v_{d_i}$, with true category embeddings, $v_{c_{i}}$ and minimises the inter-class similarity between $v_{d_i}$ and $K$ negative category embeddings. 


A key difference compared to other metric-based FSL methods is that Cat2Vec compares the distance between the query document embeddings to learned category embeddings. In contrast, FSL methods will compare the distance of query document embeddings to representations explicitly created from support set document embeddings. Thus, reasoning and classification at test time of K-shot test classes is given by equation \ref{eqn:classification}, the test class which minimises the distance ($d$) between the encoded document embedding $v_{d_i}$, and the learned category embedding of each test class where we consider $d$ as the euclidean distance, $d(v_{d_i}, v_y)  = \Vert v_{d_i} - v_y \Vert^2_2$.


\begin{equation}\label{eqn:classification}
    \argmin_{y \in Y_{\mathrm{test}}}\,\, d(v_{d_i}, v_y)
\end{equation}






\subsection{Categorical Contrastive Loss}

\begin{figure}[H]
    \centering
    \includegraphics[scale=0.45]{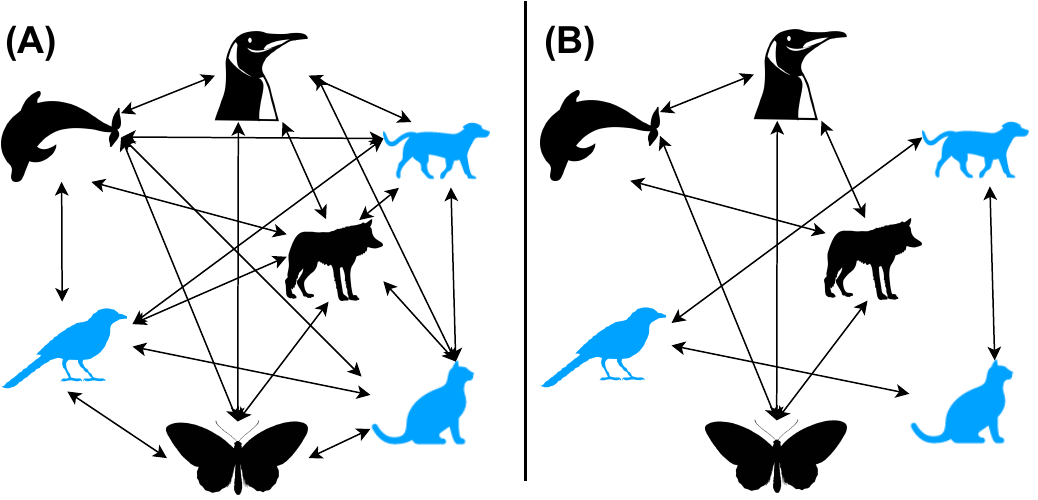}
    \caption{Possible negative pairs between K-shot test class (blue) and training classes (black) in NCE, NCA and SC loss (A) compared to CC (B).}
    \label{fig:NCEvsCC}
\end{figure}

While Cat2Vec using NCE can directly be applied to CFSL, issues arise when training using $\mathcal{D}_\mathrm{train}^\prime$ outlined in section \ref{sec:problem formulation} and contrastive based losses like NCE, NCA and SC due to treating negative pairs between training and K-shot test classes similarly. Figure \ref{fig:NCEvsCC}(A) shows the possible true classes and sampled classes involved for both document-document and document-category negatives pairs between K-shot test classes and training classes when using NCE, NCA, SC and in general, any conventional contrastive losses. For example, given two documents $d_i$, $d_2$ with corresponding classes dog and wolf, $(d_1, d_2)$ forms a possible document-document negative sample and $(d_1, \mathrm{wolf})$ forms a document-category negative sample. 

When training a model for CFSL and FSL trained using \emph{few-shot data generalisation} \cite{triantafillou2021learning} where multiple datasets are joined to create a diverse training dataset to learn more adaptive FSL models. The training classes, $\mathcal{Y}_{\mathrm{train}}$ are never explicitly used in the classification of the K-shot test classes, minimising the negative pairs component of the contrastive loss, e.g., the summation term in NCE loss, equation \ref{eqn:neg_objective}, or the numerator of SC and NCA loss, equations \ref{eqn:SC} and \ref{eqn:NCA}, can negatively impact learning and generalisation of $K$-shot test classes. Intuitively, maximising the inter-class similarity between K-shot test and training class negative pairs, e.g., dog and wolf classes in $(d_1, \mathrm{wolf})$ is unnecessary as training classes are never classified and causes the encoder and category embeddings to not learn from similar training classes as reference points for $K-shot$ classes. To solve this problem, we propose \emph{categorical contrastive loss} given by equation \ref{eqn:cat_neg_objective}, an extension of NCE loss that separates the $K$-shot test class and training classes when calculating the negative pair component of the loss during training.



\begin{small}
\begin{equation}\label{eqn:cat_neg_objective}
    \log \sigma \left(v_{c_{i}}\cdot v_{d_i}\right) + \sum^{N}_{j=1}
    \begin{dcases}
        \mathbb{E}_{n_j \sim P_\text{noise}(C_T)} \\
        \mathbb{E}_{n_j \sim \hat{P}_\text{noise}(C_K)}
    \end{dcases}
    [\log \sigma (-v_{n_j} \cdot v_{d_i})]
\end{equation}
\end{small}

In equation \ref{eqn:cat_neg_objective}, the negative pair component in NCE loss is split into two cases depending on the class of $d_i$. If $d_i$ belongs to a $K$-shot test class then categories for document-category negative pairs are only sampled from other test classes, ${n_j \sim \hat{P}_\text{noise}(C_K)}$, and similarly if $d_i$ belongs to a training classes then sampled categories can only belong to other training classes, $n_j \sim P_\text{noise}(C_T)$, as shown by figure \ref{fig:NCEvsCC}(B). In this paper, we only consider the simplest form of noise distributions where $P_{\mathrm{noise}}$ and $\hat{P}_{\mathrm{noise}}$ are uniform distributions. 

\begin{figure}[h]
  \centering \includegraphics[scale=0.35]{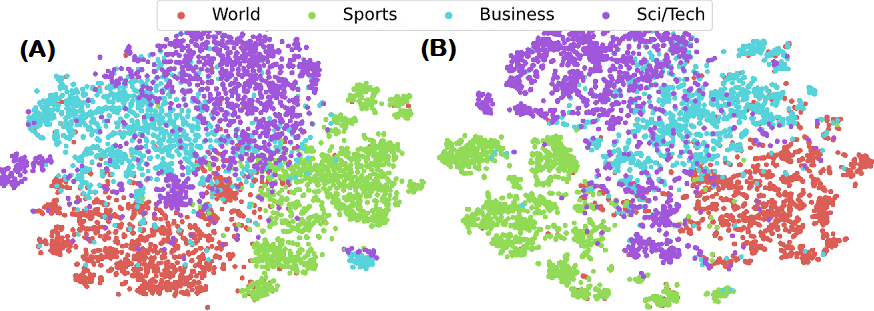}
  \caption{ TSNE NCE (A) vs CC (B) Embeddings}
 \label{fig:tsne}
\end{figure}

Figure \ref{fig:tsne} shows the TSNE of the AG dataset document embeddings from section \ref{sec:datasets} learned by Cat2Vec with a BERT encoder during CFSL using both NCA and CC losses. Figure \ref{fig:tsne}(B) using CC loss shows higher inter-class separation and clusters with high intra-class variance within each class comapred to NCE, figure \ref{fig:tsne}(A).

\section{Experiments}

The aim of our experiments is to investigate two main questions. (1) Comparison of our model against baseline models for CFSL under different tshot scenarios. (2) Explore the effect of increasing known pretrained categorical knowledge by adjusting the Wiki training set has on CFSL. 

\subsection{Datasets}\label{sec:datasets}

For training data, we create several Wikipedia datasets based on 2020 Open Wikipedia Ranking\footnote{http://wikirank-2020.di.unimi.it/} of varying sizes, $n =\ \{500, 1000, 2000, 4000\}$. The size represents the set of Wikipedia pages included based on the combined top $n$ pages for each page rank, harmonic mean and view count, where each Wiki page is its own class. 

For testing, we use predefined test splits from four standard document classification datasets previously applied to FSL, covering topic and question classification. A crucial difference compared to episodic training is that we do not need to partition the dataset based on classes for training and use all classes for K-shot testing.

The datasets fall into four distinct scenarios in relation to the training Wiki datasets and experimentation using these datasets covers a broad spectrum of scenarios for CFSL and FSL. \textbf{(C1)} \textbf{DBpedia} (topic classification of Wikipedia articles): Number of test classes is relatively large, training and test class distributions are similar with low test class inter-class similarity. \textbf{(C2)} \textbf{AG News} (topic classification of news articles from various sources): Number of test classes is relatively small, and differences between training and test classes are distinct with low test class inter-class similarity. \textbf{(C3)} \textbf{20News} (topic classification of news articles from various sources): Number of test classes are relatively large, difference between training and test classes are distinct with high test class inter-class similarity, e.g., { \tt comp.sys.ibm.pc.hardware} and { \tt comp.sys.mac.hardware}. \textbf{(C4)} \textbf{Yahoo} (question classification of questions with best responses): Moderate number of test classes and differences between training and test classes are distinct with low test class inter-class similarity.


%

\begin{table}[H]
\centering
\resizebox{\columnwidth}{!}{%
\begin{tabular}{|c|c|c|c|}\hline
 Dataset & Classes & Documents & Average Words \\\hline
 Wiki500 & 1085 & 1085 & 7658.50 \\
 Wiki1000 & 2124 & 2124 & 6866.16 \\
 Wiki2000 & 4198 & 4198 & 6123.61 \\
 Wiki4000 & 8088 & 8088 & 5412.84 \\
 \hline
 DBPedia & 14 & 70000 & 48.88 \\
  AG  & 4 & 7600 & 36.91 \\ 
   20News & 20 & 17843 & 276.11 \\
 Yahoo & 10 & 60000 & 93.13 \\ \hline
\end{tabular}
}
\caption{Dataset Statistics.}
\end{table}

\begin{table*}[t]
\centering
\resizebox{\columnwidth * 2}{!}{%
\begin{tabular}{|c|c c c|c c c|c c c|c c c|}\hline
 & \multicolumn{3}{c|}{DBPedia} & \multicolumn{3}{c|}{AG News}  & \multicolumn{3}{c|}{20News} & \multicolumn{3}{c|}{Yahoo} \\ 
 &  1-shot & 5-shot & 20-shot &  1-shot & 5-shot & 20-shot &  1-shot & 5-shot & 20-shot &  1-shot & 5-shot & 20-shot \\\hline

Bi-LSTM (xEnt)  & 40.32 & 69.36 & 83.44 & 29.49 & 60.18 & 65.13 & 10.56 & 18.77 & 25.26 & 16.41 & 31.50 & 42.14\\
Bi-LSTM (Distil)  & 43.10 & 73.97 & 86.92 & 32.44 & 58.26 & 66.25 & 11.51 & 18.22 & 34.73 & 15.42 & 33.04 & 44.96 \\
Bi-LSTM (NCE) & 39.47 & 74.98 & 83.31 & 36.46 & 65.57 & 62.63 & \textbf{19.34} & 24.25 & 32.19 & 18.32 & 39.51 & 40.70  \\ 
\hline
Cat2Vec$_{\mathrm{Bi-LSTM}}$ & 43.91 & 75.84 & 83.70 & 42.14 & 71.07 & 78.98 & 17.42 & 29.73 & 53.01  & 18.98 & 40.29 & 48.83\\ \hline

BERT (xEnt)  & 39.77 & 76.17 & 93.90  &  43.89 &  70.46 & 82.93 & 11.51  & 23.10 & 46.53 & 18.33 & 33.23 & 56.02  \\
BERT (Distil)  & 41.43 & 77.22 & 93.30 & 41.26 & 72.68 & 81.10 & 12.31 & 20.85 & 45.71 & 23.98 & 33.05 & 56.16 \\
BERT (NCE) & 46.78 & 85.59 & 68.73 & 49.47 & 72.58 & 61.25 & 11.27 & 22.80 & 18.30 & 30.83 & 36.98 & 19.37 \\
\hline
Cat2Vec$_{\mathrm{BERT}}$  & \textbf{56.24} & \textbf{90.17} & \textbf{96.84} & \textbf{60.25} & \textbf{82.48} & \textbf{84.68}
 & 15.56 & \textbf{41.39}  & \textbf{54.62} & \textbf{37.16}  & \textbf{51.09} & \textbf{58.90} \\\hline


\end{tabular}
}
\caption{1, 5 and 20-Shot 1-TShot Constrained Few-Shot Learning Results.} \label{table:cfsl}
\end{table*}

\subsection{Benchmark Algorithms}

In this paper, we select two popular NLP encoding methods based on pretrained LMs.
\textbf{Bi-LSTM}: Single-layered Bi-LSTM using attention with context \cite{yang2016hierarchical} and pretrained Wikipedia Word2Vec embeddings.
\textbf{BERT}: Pretrained transformer based model \cite{devlin2018bert}, where pretrained BERT-base-uncased model is fine-tuned to perform document classification/FSL.

Using these pretrained LMs as encoders, we train 6 baseline models: (\textbf{xEnt}) Standard softmax output and cross-entropy loss (\textbf{Distil}) xEnt model with self knowledge distillation \cite{tian2020rethinking} (\textbf{NCE}) Categorical negative sampling with NCE loss as described in section \ref{sec:cat2vecCFSL} (\textbf{SC}) Projection network and standard softmax output using SC loss \cite{laenen2021episodes} (\textbf{NCA}) Projection network and standard softmax output using NCA loss \cite{laenen2021episodes}  (\textbf{Cat2Vec}) Cat2Vec using our proposed CC loss.

\subsection{Evaluation Metrics}

Due to the nature of episodic and task-based training, the evaluation method often obscures the true performance of a $K$-shot learning model across the whole dataset. Typically, the reported accuracy of episodic and task-based FSL model is the average accuracy across randomly sampled test episodes, ($E = \{(Q_1, S_1), (Q_2, S_2), \ldots, (Q_n, S_n)\}$), given by equation \ref{eqn:episode_evalaution} where $y_q$ is the true class label of document $q$, $Q$, $S$ are query, support sets and $Q_E$ is the total number of query examples across all test episodes. Because FSL is highly sensitive to the choice of support examples, the accuracy derived by equation \ref{eqn:episode_evalaution} obscures the true performance of the model as sampled query sets usually contains a small subset of all test instance and does not show the impact of each support set across the entire dataset.


\begin{equation}\label{eqn:episode_evalaution}
    \frac{1}{Q_E}\sum_{Q,S \in E}\sum_{q \in Q}{\mathrm{Pred}(q, S) = y_q}
\end{equation}

To more extensively gauge the significance of each sampled support set on the accuracy, we take a more traditional approach to calculating the accuracy by reporting the average accuracy of each support set against all test data, given by equation \ref{eqn:our_evalaution} where $X$ is the set of test documents.

\begin{equation}\label{eqn:our_evalaution}
    \frac{1}{|X|}\sum_{x \in X}{\mathrm{Pred}(x, S) = y_x}
\end{equation}

\section{Experiment Results}

Table \ref{table:cfsl} displays the average results across three different sampled support sets for each baseline using the Wiki1000 as the training dataset and the first 50 words of each document as input. In general, Cat2Vec$_{\mathrm{BERT}}$ with CC loss achieves state-of-the-art performance in almost all cases compared to baselines across 1,5 and 20-shot 1-Tshot CFSL.

Our model does particularly well in the 5-shot case, a typically demanding case where we achieve a significant 4.58\%, 9.8\%, 17.14\% and 11.57\% increase in accuracy compared to the next best baseline on the DPPedia, AG, 20News and Yahoo datasets. The two biggest increases come from the 20News and Yahoo datasets relating to C3 and C4, indicating our method improves performance under challenging cases with a large number of classes and differences between test and training distributions. Also, as $K$-shots increase, we observe diminishing returns in terms of performance that are consistent with the trends observed in FSL.

In the 1-shot case where our model was beaten by the Bi-LSTM (NCE) baseline, because of the high inter-class similarity in the 20News dataset and only having one example of each class is not sufficient information to learn inter-class differences when there is a significant difference between training and test distributions.



\subsection{Comparison to Conventional Classifiers}

As our approach to CFSL allows us to evaluate on all classes, albeit with distribution shifts between training and test data, this allows us to indirectly compare CFSL models to conventionally trained classifiers that take advantage of large amounts of labelled data for each class during training. Table \ref{table:convential} shows the results of applying standard xEnt variant of each pre-trained LM having access to up to 1.4 million labelled examples from the training split outlined in table \ref{table:conventionaldataset}.

\begin{table}[H]
\centering 
\scalebox{0.85}{
\begin{tabular}{|c|c|c|}\hline
 Dataset & Training Instances  & Test Instances \\\hline
 DBPedia & 560000 & 70000 \\
 AG & 120000 & 7600 \\
 20News & 14274 & 3569  \\
 Yahoo & 1400000 & 60000  \\\hline

\end{tabular}
}
\caption{Conventional Classifier Dataset Instances.} \label{table:conventionaldataset}
\end{table}

\vspace{-0.5cm}

\begin{table}[H]
\centering 
\resizebox{\columnwidth}{!}{%
\begin{tabular}{|c|c|c|c|c|}\hline
 Model & DBPedia  & AG & 20News & Yahoo \\\hline
 Bi-LSTM & 98.58 & 88.71 & 68.13 & 70.74 \\
 BERT & 99.05 & 93.35 & 88.55 & 73.83 \\
  \hline
\end{tabular}
}
\caption{Conventional Classifier Results.} \label{table:convential}
\end{table}
\vspace{-0.3cm}

In general, the results indicate the 20News and Yahoo datasets are more challenging as both conventional and CFSL have the lowest performance, with the most considerable difference being on the 20News dataset, where CFSL has trouble learning from few examples when there is high inter-class similarity. Further comparisons indicate that our Cat2Vec$_{\mathrm{BERT}}$ CFSL model can be competitive with conventional classifiers under certain conditions while using significantly less labelled data, e.g., in C1, where the distribution of the CFSL training set is similar to the test set with an 8.88\% and 2.21\% difference in the 5-tshot and 20-tshot and to a lesser extent in C2, where the relative number of classes is small with a 12.26\% and 8.9\% difference.

\subsection{Increasing Training Shots}

As SC and NCA losses are not calculable for 1-tshot due to the lack of multiple positive samples for each class. We experiment with a 5-tshot scenario to compare our method against these baselines and explore the role of additional training data in CFSL. To construct such a scenario, we partition each document in the Wiki1000 training sets into partitions of text size 50 (the text input size of all models) and use the first five as the tshot documents. Table \ref{table:5tshot} shows the average 5-shot 5-tshot results from the same three support sets where we observe a general increase in accuracy across all models on Yahoo and 20News datasets compared to 1-tshot results as all models benefit from additional data. Overall, our Cat2Vec$_{\mathrm{BERT}}$ outperforms all models for 5-tshot as well. 

\begin{table}[H]
\centering
\resizebox{\columnwidth}{!}{%
\begin{tabular}{|c|c|c|c|c|}\hline
Model & DBPedia &  AG & 20News & Yahoo \\\hline
Bi-LSTM (xEnt)& 76.28 & 56.02  & 27.07  & 29.50 \\
Bi-LSTM (Distil) & 78.32 & 60.47  & 26.90  & 37.87 \\
Bi-LSTM (NCE)  & 68.00 & 68.77  & 40.36 & 40.41  \\ 
Bi-LSTM (SC)  & 68.85 & 60.63  & 21.16 & 43.40
  \\ 
Bi-LSTM (NCA)  & 84.04 & 67.86  & 32.01  & 42.28 \\ \hline
Cat2Vec$_{\mathrm{Bi-LSTM}}$  & 69.76 & 71.31 & 40.92  & 41.36\\ \hline
BERT (xEnt) & 72.25 & 66.61  & 33.96 & 43.75
\\
BERT (Distil) & 73.22 & 68.85  &  33.41 & 44.62 \\
BERT (NCE)  & 84.36  & 80.67 &  38.43  &  54.95 \\ 
BERT (SC)  & 72.94 & 70.26 & 38.45  & 40.71 \\ 
BERT (NCA)  & 82.11 & 69.36  & 39.67   & 46.27 \\ \hline
Cat2Vec$_{\mathrm{BERT}}$  & \textbf{85.40} & \textbf{82.16}
 & \textbf{47.15}  & \textbf{56.41} \\
  \hline
\end{tabular}
}
\caption{5-TShot 5-Shot CFSL Results} \label{table:5tshot}
\end{table}

To allow a FSL using episodic training to train using the Wiki1000 dataset, we similarly partition the training set documents into 50 text size partitions with the same class labels. While this method is artificial, it allows us to benchmark existing FSL methods using the training Wiki1000 dataset to understand how current FSL methods handle distribution shifts between the training and test data and evaluate FSL based on the metric outlined in equation \ref{eqn:our_evalaution}. 
Table \ref{table:FSL_wiki} displays the average result using the same 3 sampled support sets on each test dataset when each model was trained on all Wiki1000 partitions. The results indicate that our 5-shot 1-tshot or 5-tshot Cat2Vec$_{\mathrm{BERT}}$ models from tables \ref{table:cfsl} and \ref{table:5tshot} indirectly outperforms all FSL models in the sub-optimal FSL case where there is a shift between test and training distributions.




\begin{table}[H]
\centering
\resizebox{\columnwidth}{!}{%
\begin{tabular}{|c|c |c |c | c|}\hline
Model & \multicolumn{1}{c|}{DBPedia} & \multicolumn{1}{c|}{AG}  & \multicolumn{1}{c|}{20News} & \multicolumn{1}{c|}{Yahoo} \\ \hline

Proto  [\citenum{snell2017prototypical}] & 62.54 & 67.29 & 36.12  & 34.38 \\ 
Induction  [\citenum{geng2019induction}]   & 66.40 & 74.30   &  34.04 & 36.47 \\ 
Meta (R2D2) [\citenum{bertinetto2018meta}]   & 85.61 & 77.13   & 43.46 & 38.76 \\ \hline
\end{tabular}
}
\caption{FSL with Training Distribution Shift.} \label{table:FSL_wiki}
\end{table}

\subsection{Increasing Categorical Knowledge}

To determine the impact of categorical knowledge and how much categorical is useful, we experiment with using Wiki500 to Wiki4000 as training datasets which explicitly alters the knowledge known from 1085 to 8088 classes. Hypothetically, having access to more diverse training data is beneficial for learning precise categorical test class embeddings at the expense of longer training times.

\begin{figure}[h]
  \centering \includegraphics[scale=0.48]{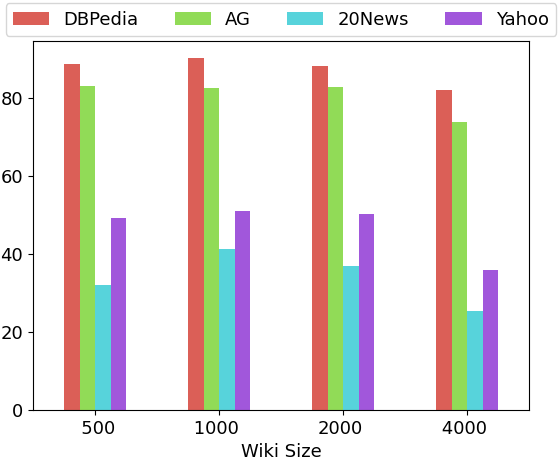}
  \caption{Cat2Vec: Effects of Training Knowledge.}
 \label{fig:increasing wiki}
\end{figure}

Figure \ref{fig:increasing wiki} shows the average results across three sampled support sets for 5-shot 1-tshot CFSL using Cat2Vec$_{\mathrm{BERT}}$ for each Wiki training dataset. The results indicate there is not much difference when using Wiki500-2000 datasets with minor improvements on the 20News dataset from the extra knowledge. However, after some inflection point, Wiki4000, having access to too much knowledge can be detrimental leading to a significant drop in accuracy up to 15\% as it becomes difficult to learn the relationships between additional classes which do not add value. Additional knowledge is useful when it follows a similar distribution to the $K$-shot test class distribution or is diverse from the original training classes, e.g., joining different dataset together as in few-shot data generalisation \cite{triantafillou2021learning}. This mimics the cognitive phenomenon known as \emph{information overload}, the presence of excess information causing indecision or uncertainty in decision making.

\subsection{Discussion and Error Analysis}

Results in tables \ref{table:cfsl} and \ref{table:5tshot} indicate several challenges of CFSL and in general FSL. These challenges can be described in terms of the four cases based on the characteristics of each dataset outlined in section \ref{sec:datasets}. In general, under the conditions that the distributions of the training and $K$-shot test class data are similar and there is low similarity between test classes, CFSL and FSL perform the best, e.g., C1 Wiki1000 and DBPedia. This is evident from all results in tables \ref{table:cfsl} and \ref{table:5tshot} where all models achieve the highest accuracy on the DBPedia dataset. However, when distributions of training and $K$-shot test class data differ significantly, e.g., C2, C3 and C4 are when deficiencies in CFSL and FSL approaches become more obvious.

In C2, where the number of test classes is relatively small and the similarities between test classes are low, e.g., AG news, CFSL and FSL tend to perform well. However, compared to conventional classifiers, the CFSL and FSL methods are limited by the distribution shifts between the training and test and to a lesser extent, the low number of labelled test examples resulting in the 11.77\% difference in accuracy between the best models.

In C3, where number of test classes are relatively large and high inter-class similarity between some test classes, e.g. 20News, CFSL and FSL struggle to differentiate between similar classes based on only a few examples due to the high similarity between texts between classes. In this case, our evaluation metric shows that the choice of documents for the support set is critical. This is non-trivial as the $K$ examples of each class should diversely represent the distribution of each class while minimising the similarity between documents of other classes. To our knowledge, no work researches the significance and choice of the support set in the context of FSL. In comparison to conventional classifiers and CFSL and FSL classifiers struggle the most with a 41\% difference in accuracy between the best models.
Finally, in C4, where the number of test classes are relatively large with low similarity between test classes, e.g. Yahoo! Answers, as the number of test classes increases the difficulty for a CFSL or FSL model to learn and differentiate between the uniqueness of each class increase. Especially when there are overarching similarities between classes, such as the structure of the text in the form of questions and classes cover very broad and potentially overlapping topics. Classifying such a dataset is difficult even for conventional classifiers with access to many labelled examples of each class, as shown in table \ref{table:convential}.

\section{Conclusion and Future Work}

In this paper, we introduce the novel but extremely challenging sub-problem of few-shot learning, constrained few-shot learning that inherently promotes learning from diverse but sparse datasets. We propose that CFSL can be performed using Cat2Vec with a novel categorical contrastive loss on CFSL document classification. Our evaluation method is consistent with standard classification to not obscure FSL and CFSL methods' performance and allows us to indirectly compare across conventional, FSL and CFSL methods. Our experiments show that our method achieves state-of-the-art performances compared to all baselines in various CFSL settings. Future work includes improving the performance of CFSL on datasets that exhibit high inter-class similarity and improving the adaptive capabilities when there are considerable differences between the training and test distributions.

\bibliography{cftt}







\end{document}